\documentclass[11pt]{article}

\usepackage[preprint]{acl}
\usepackage{times}
\usepackage{latexsym}
\usepackage[T1]{fontenc}
\usepackage[utf8]{inputenc}
\usepackage{microtype}
\usepackage{inconsolata}
\usepackage{graphicx}
\usepackage{placeins}
\usepackage{booktabs}
\usepackage{hanging}
\usepackage{multicol}
\usepackage{listings}

\title{Semantic Variability of LLM-Generated Replies Across LLMs: Implications for Designing Conversation-Based Assessment}
\author{
  Jiangang Hao\\
  ETS Research Institute\\
  Princeton, New Jersey, USA\\
  \texttt{jhao@ets.org}\\
  \small Accepted to AI in Measurement and Education Conference -- AIME 2026
}
\date{}

\begin{document}
\raggedbottom
\maketitle

\begin{abstract}
This study examines whether LLM-generated replies remain semantically consistent when the underlying LLM changes. Using messages from real collaborative conversations, we compared the semantic similarity of generated replies across LLMs under two conditions: with and without preceding chat history. Results show that model choice and conversational context both affect response similarity and alignment with human replies. These findings indicate that prompting and conversational context alone may not be sufficient to preserve response consistency across LLMs, highlighting the need for infrastructure and design strategies that can maintain stable and comparable responses amid the rapid and continuous evolution of LLMs.
\end{abstract}

\section{Introduction}

Large language models (LLMs) and generative AI are creating a new paradigm for the assessment of complex skills such as communication and collaboration (Hao et al., 2024). These skills have long been difficult to measure at scale using conventional item formats as they are manifested through situative and dynamic interactions rather than isolated responses (Mislevy, 2018). LLM-based agents can engage learners in natural, adaptive interactions, creating new opportunities to elicit richer and more authentic evidence of these skills.

At the same time, LLM-enabled interactions introduce a fundamental tension between adaptivity and standardization. Standardized assessment assumes that comparable learner inputs should lead to comparable assessment conditions and opportunities to demonstrate the evidence of target construct. In contrast, LLM-based agents generate responses dynamically, adapting to learner input, conversational context, and model behavior. As a result, different learners may experience different interaction pathways, and the same learner input may receive semantically different replies depending on the LLM, prompt, deployment setting, or prior chat history. This challenge is further complicated by the rapid evolution of LLMs, as new models are frequently introduced and older models are updated or retired, making it difficult to preserve stable assessment conditions over time.

From a measurement perspective, the key issue is not whether AI responses vary, but whether they remain sufficiently consistent for measuring the intended constructs. AI-generated replies need not be identical in wording, but they should preserve the intended assessment function, sustain task-relevant interaction patterns, and provide comparable opportunities for learners to demonstrate the target construct. Otherwise, uncontrolled variability may introduce construct-irrelevant variance that threatens validity, reliability, fairness, and comparability, which are the foundational requirements of educational assessment (AERA, APA, \& NCME, 2014).

This challenge is not entirely new. Assessment has long dealt with variability arising from task contexts, raters, and interaction processes while seeking to support comparable score interpretations. In human-led interviews, for example, interviewers may differ in speaking style, follow-up prompts, or interpersonal dynamics, yet the assessment is expected to yield interpretable evaluations of the target construct. Similarly, standardized writing assessments may use different prompts across forms or administrations while maintaining score comparability. Thus, the goal is not to eliminate variability, but to ensure that it remains within acceptable bounds for the intended use and does not introduce construct-irrelevant variance that undermines validity, reliability, and fairness.

Traditional conversational systems, such as those used in the intelligent tutoring systems, typically employ modular architectures consisting of intent detection, dialogue state tracking, and rule-based response selection (Corbett et al., 1997; Graesser et al., 2018; Young et al., 2013; Zapata-Rivera et al., 2015). While effective for well-defined interactions, such rule-based and symbolic approaches often lack flexibility in handling open-ended conversations and are difficult to scale. In contrast, LLM-based systems rely primarily on prompts and conversation history to guide response generation, replacing much of the traditional dialogue management pipeline with natural language instructions (Wei et al., 2022), which enables more natural interactions but also results in greater variability in generated responses.

To optimize this tradeoff, LLM-symbolic, or more broadly neuro-symbolic, approaches are increasingly adopted in conversation-based learning and assessment tasks (e.g., Hou et al., 2025; Zapata-Rivera et al., 2026). In such systems, LLMs are used to interpret learner input, detect intent, classify dialogue moves, or generate responses, while rule-based actions are executed either by a separate software module or by an LLM prompted to follow predefined rules. In principle, such a design can preserve the flexibility of LLMs while maintaining greater control over the consistency of relevant responses.

However, in practice, even with a neuro-symbolic design, practical design decisions depend heavily on two LLM capabilities: accurately classifying utterances and generating semantically consistent responses in similar conversational contexts. Hao et al. (2025, 2026) have demonstrated the accuracy of LLM-based utterance coding; therefore, the present study focuses on the second capability. Because LLMs evolve rapidly, with frequent updates, new model releases, and model retirements, understanding how model choice and conversational context affect the semantic consistency of LLM-generated responses is a crucial step toward building robust conversation-based assessments.

In this study, we examine the semantic variability of LLM-generated replies to messages drawn from real collaborative conversations across different LLMs and conversational context conditions. By analyzing within-model similarity, alignment with human replies, and cross-model similarity, we provide empirical evidence on how changes in the underlying LLM may affect response consistency. The results highlight the need for infrastructure and design strategies that can support consistent and comparable conversation-based assessment as LLMs continue to evolve.

\section{Methods}

\subsection{Data Source and Sampling}

This study used conversational data from dyadic online collaborative problem-solving science tasks. The data were collected through Amazon Mechanical Turk in 2018. In each task, two participants worked together through chat-based communication, producing approximately 50 to 100 turns per team. To characterize the progression of the interaction, messages were divided into early, middle, and late segments based on their relative turn positions within each conversation.

For the present study, we randomly selected data from 99 teams. Each message was treated as a potential focal message, and the actual reply produced by the human teammate in the original interaction was retained as the reference human response. Human replies were coded as having low, medium, or high relevance to the corresponding focal message. For each focal message, all preceding communication turns were used as the available conversational history.

The analysis focused on late-stage focal messages whose corresponding human replies were coded as highly relevant. This selection resulted in 61 focal messages. These messages were chosen because they were supported by richer prior conversational context and had clearly interpretable human replies, making them suitable cases for examining the semantic consistency of LLM-generated responses.

\subsection{Experimental Conditions}

The study evaluated four LLMs: GPT-4o mini, GPT-5.4, GPT-5.4 mini, and GPT-5.4 nano. All models were prompted using the same system instruction, shown in Appendix A. To examine the role of conversational context, we compared two prompting conditions. In the No History condition, the prompt included only the system instruction and the focal message. In the With History condition, the prompt additionally included all preceding conversational turns as context. For each focal message under each condition, each LLM generated 100 independent responses using the same prompt. When calling an LLM API, temperature is a decoding parameter that affects the randomness of response generation. Because temperature can influence response variability, we set it to 0.7 when supported by the model deployment endpoint. Among the models evaluated, this option was available only for GPT-4o mini.

\subsection{Semantic Similarity}

Semantic similarity was quantified using the cosine similarity of text embeddings, a widely used approach in natural language processing for comparing the meanings of texts. First, each focal message, human reply, and LLM-generated reply was encoded as an embedding vector using OpenAI's text-embedding-3-large model. The embedding model represents each text in a high-dimensional semantic space, where semantically similar texts are located closer together. Cosine similarity was then computed between pairs of embedding vectors to measure the degree of semantic similarity between texts, with higher values indicating greater semantic alignment.

\subsection{Exploratory Analyses}

First: We compared the mean pairwise similarity among LLM-generated replies across LLMs and chat history conditions for each of the focal messages. This analysis intended to evaluate whether different models and prompting conditions produced semantically similar response sets for the same focal messages. We also computed the mean similarity between each LLM-generated reply and the corresponding human reply to examine how closely LLM responses aligned with human responses.

Second: We compared the similarity among replies generated by the same LLM with the similarity among replies generated by different LLMs for the same focal message under each chat history condition. This analysis was intended to examine whether replies remain semantically comparable when the generating LLM changes.

Third: We examined whether semantically similar focal messages elicited semantically similar LLM replies. For each pair of focal messages, we computed the cosine similarity between the focal-message embeddings. We then compared the generated reply sets across the two focal messages. This analysis tested whether LLMs generated comparable replies when the input focal messages were semantically similar.

Fourth: We directly compared the semantic similarity of LLM-generated replies produced with and without chat history. This analysis examined the extent to which including prior conversational context changed the semantic content of the generated replies.

\subsection{Statistical Analysis}

In addition to the exploratory analyses, we conducted statistical analyses to estimate overall effects. We fitted linear mixed-effects models to examine whether LLM type and chat history condition were associated with two dimensions of generated-reply similarity. The first outcome was the mean pairwise similarity among generated replies, reflecting average semantic consistency. The second outcome was the standard deviation of pairwise similarity, reflecting the spread or variability of semantic similarity. These outcomes were modeled separately because they capture distinct aspects of response behavior: higher mean similarity indicates greater semantic consistency, whereas lower variability indicates more stable response generation.

The fixed effects included LLM type, chat history condition, their interaction, and the mean similarity between LLM-generated replies and the observed human reply. GPT-4o mini and the No History condition served as the reference categories. A random intercept for focal message was included to account for repeated observations of the same focal message across models and chat history conditions.

\section{Results and Findings}

\subsection{Exploratory Analysis 1 Results}

Figure 1 compares two forms of semantic similarity across LLMs and context conditions. The top panel shows the distribution of mean pairwise similarity among the 100 LLM-generated replies for the same focal message. The bottom panel shows the distribution of mean similarity between the LLM-generated replies and human replies.

These results indicate that including chat history does not make LLM replies more similar to one another in all cases. Its effect on the similarity depends on LLMs. The LLM-generated replies are consistently not similar with human replies, though including chat history can slightly improve the situation.

\begin{figure*}[!tbp]
\centering
\includegraphics[width=0.76\textwidth]{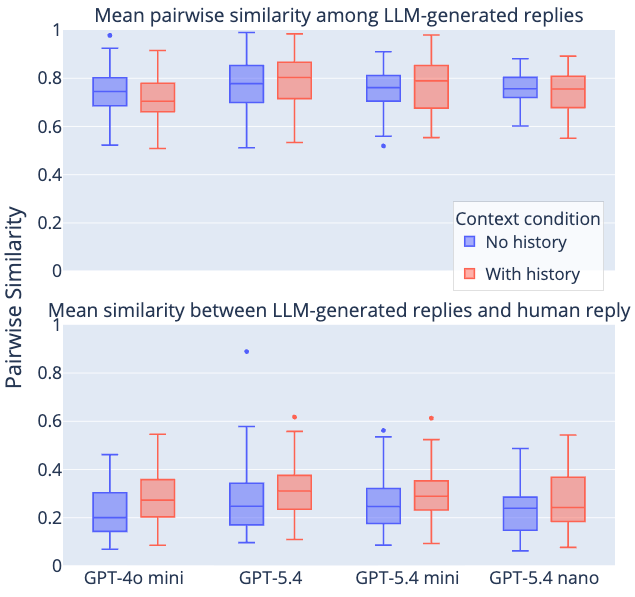}
\caption{Semantic similarity by model and context condition.}
\label{fig:1}
\end{figure*}

\subsection{Exploratory Analysis 2 Results}

Figure 2 compares the average similarity of replies generated by the same and different LLMs for the same focal messages. The diagonal cells show mean pairwise similarity among replies generated by the same LLM, while the off-diagonal cells show mean similarity between replies generated by different LLMs. Separate heatmaps are shown for the no-history and with-history conditions.

The diagonal values were consistently higher than the off-diagonal values in both conditions. This indicates that each LLM generated replies that were more similar to its own responses than to responses generated by other LLMs. In the no-history condition, within-LLM similarity ranged from 0.751 to 0.780, whereas between-LLM similarity ranged from 0.443 to 0.595. In the with-history condition, within-LLM similarity ranged from 0.715 to 0.795, while between-LLM similarity ranged from 0.475 to 0.604.

Among the models evaluated, GPT-5.4 exhibited the highest within-model similarity under both prompting conditions, particularly when conversational history was provided. Cross-model comparisons further showed that models within the GPT-5.4 family were more similar to one another than to GPT-4o mini. Together, these findings suggest that response consistency is influenced by model architecture and lineage, with models from the same family producing more semantically comparable replies than models from different families or generations.

\begin{figure*}[!tbp]
\centering
\includegraphics[width=0.72\textwidth]{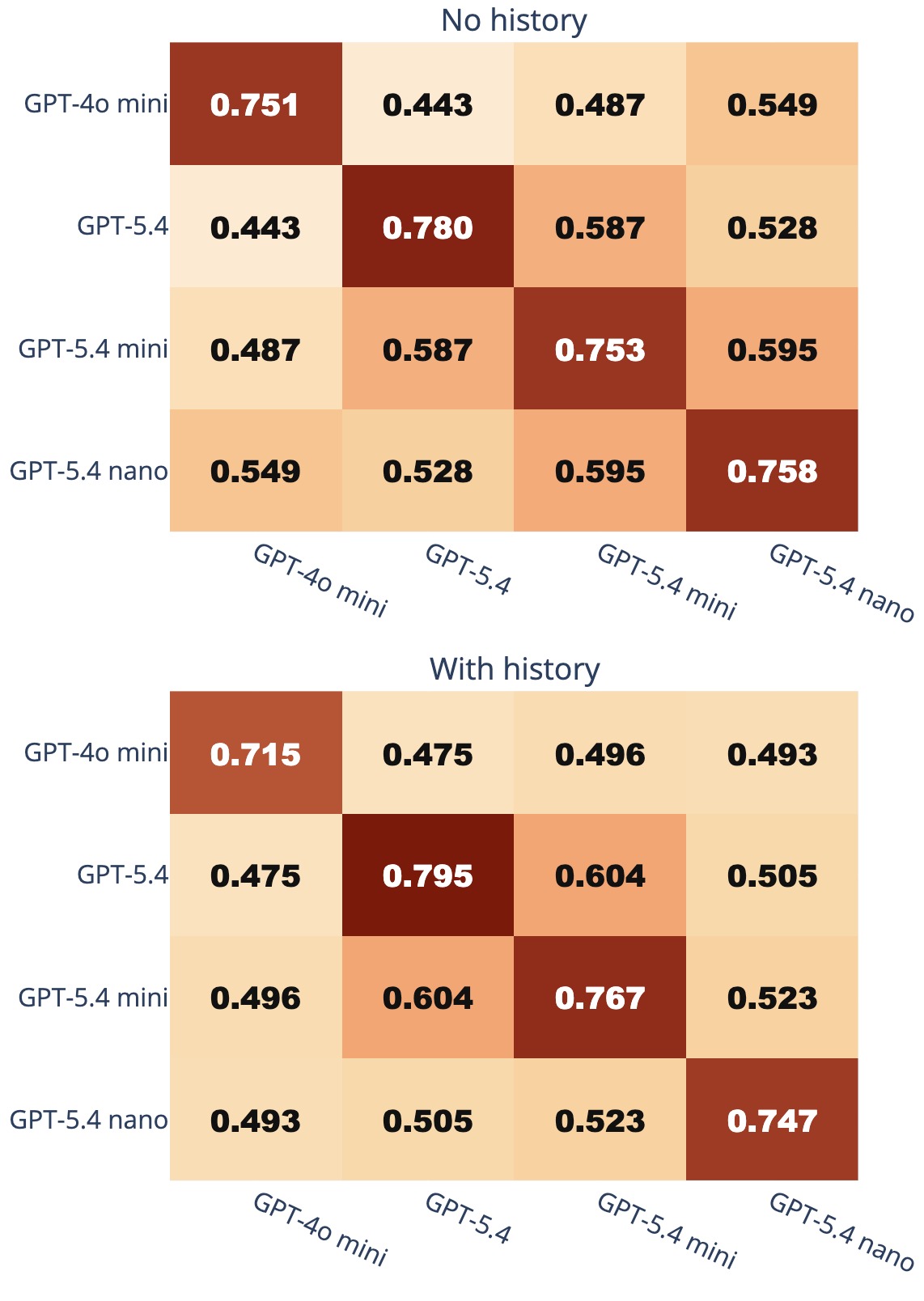}
\caption{Average within- and between-LLM similarity by context condition.}
\label{fig:2}
\end{figure*}

\subsection{Exploratory Analysis 3 Results}

Figure 3 examines whether LLMs generated comparable replies for semantically similar focal messages. The x-axis represents the mean similarity between pairs of focal messages within each bin, and the y-axis represents the mean similarity between the LLM replies generated for those focal-message pairs. Each panel corresponds to one LLM, and separate lines are shown for the no-history and with-history conditions.

Across all four LLMs, the no-history condition showed a clear positive relationship between the similarity of focal-messages and that of LLM-generated replies. This means that when two focal messages were more semantically similar, the replies generated by the LLMs were also more similar. This pattern suggests that, without additional conversational context, LLM responses were strongly anchored to the focal message itself.

The with-history condition also showed a positive correlation but was generally weaker. In several models, especially GPT-5.4 mini and GPT-5.4 nano, the with-history line increased more slowly and remained below the no-history line at higher levels of focal-message similarity. This suggests that adding chat history introduced additional contextual information that made replies less dependent on focal-message similarity alone.

\begin{figure*}[!tbp]
\centering
\includegraphics[width=0.68\textwidth]{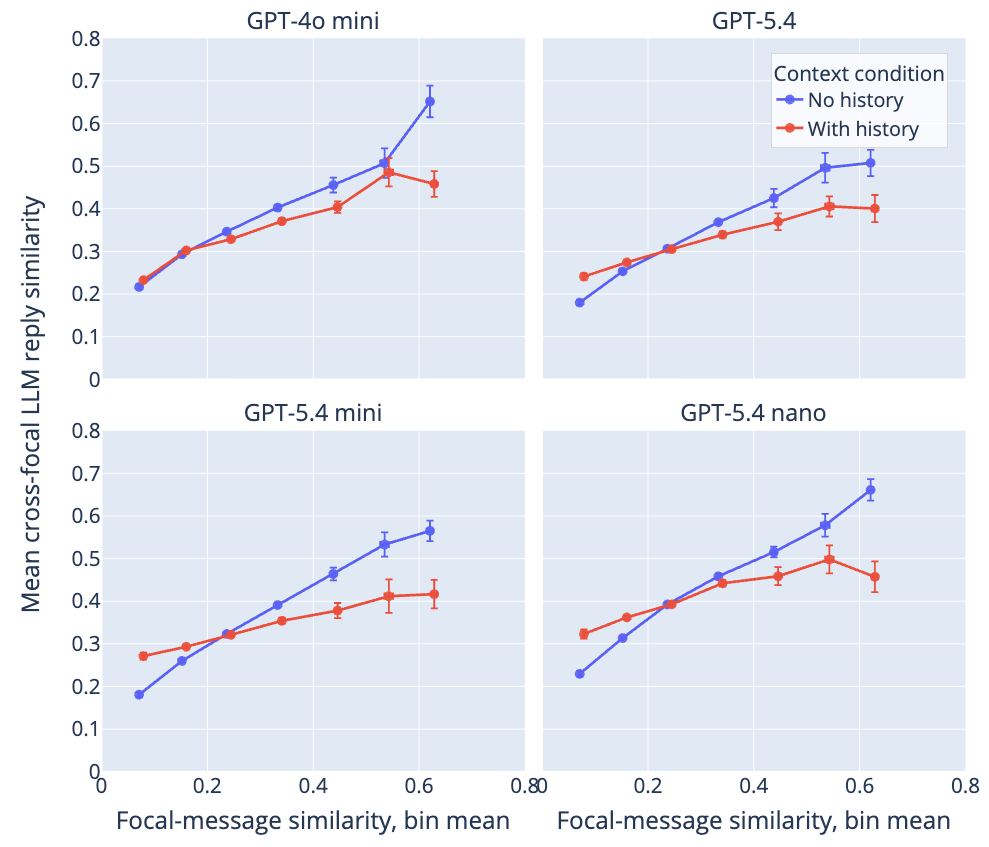}
\caption{Relationship between similarity of focal messages and that of LLM generated replies by model and context condition.}
\label{fig:3}
\end{figure*}

\subsection{Exploratory Analysis 4 Results}

Figure 4 examines how similar each model's with-history replies were to its no-history replies for the same focal messages. For each focal message and LLM, we compared the 100 replies generated without chat history to the 100 replies generated with chat history. This produced a mean with-history/no-history reply similarity for each focal message and model.

The distributions were broadly similar across the four LLMs. Median cross-history similarity was around 0.40 to 0.45 for all models, indicating that replies generated with history were only moderately similar to replies generated without history. This suggests that adding chat history meaningfully changed the semantic content of the generated replies.

\begin{figure*}[!tbp]
\centering
\includegraphics[width=0.76\textwidth]{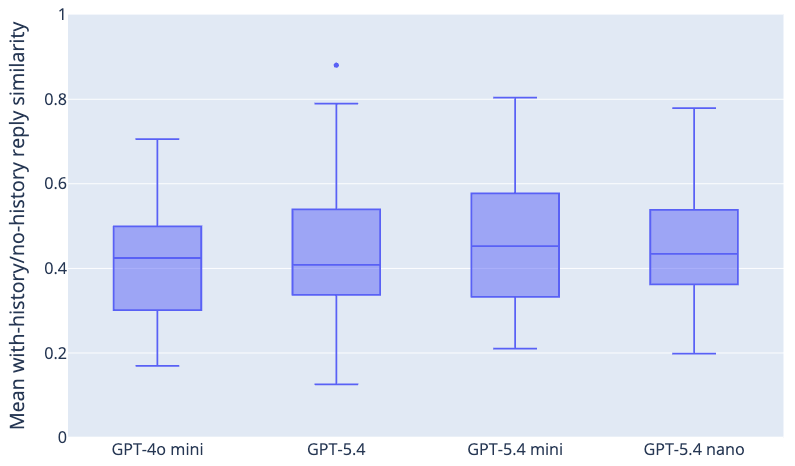}
\caption{Distribution of similarity from messages generated with and without chat history.}
\label{fig:4}
\end{figure*}

\subsection{Statistical Analysis Results}

Tables 1 and 2 report the mixed-effects modeling results for two outcomes: the mean pairwise similarity among LLM-generated replies and the variability (standard deviation) of those pairwise similarities, respectively. These models tested whether these outcomes differed by LLM model, history condition, and their interaction, while accounting for repeated observations from the same focal message. The models also included LLM-human similarity as a covariate.

For the Similarity Mean Model, the with-history condition was associated with lower mean pairwise similarity for the reference model, GPT-4o mini. However, the interaction terms for GPT-5.4 and GPT-5.4 mini were positive and statistically significant. This suggests that the effect of chat history differed across models. Compared with GPT-4o mini, GPT-5.4 and GPT-5.4 mini were less negatively affected by the inclusion of history, and GPT-5.4 showed a stronger tendency toward higher mean similarity when history was included. In addition, LLM-human similarity was positively associated with mean pairwise similarity, indicating that replies closer to the human response also tended to be more similar to one another.

For the Similarity Variation Model, GPT-5.4 showed higher variation in pairwise similarity than GPT-4o mini, whereas GPT-5.4 nano showed significantly lower variation. The with-history condition increased variation for the reference model, GPT-4o mini. The interaction terms for the other GPT-5.4 models were negative, although not all were statistically significant, suggesting that the increase in variation under history was smaller for these models. LLM-human similarity was negatively associated with variation, meaning that replies more similar to the human response tended to show less variability in their pairwise similarity.

\begin{table*}[t]
\centering
\small
\caption{Mixed-effects results for mean pairwise similarity among LLM-generated replies. Reference categories are GPT-4o mini and No History. Significance codes: *** $p<.001$, ** $p<.01$, * $p<.05$, . $p<.10$.}
\label{tab:similarity-mean-model}
\begin{tabular}{lrrrrc}
\hline
Term & Estimate & Std. Error & $z$ & $p$ & Sig. \\
\hline
(Intercept) & 0.7267 & 0.0150 & 48.485 & $<.001$ & *** \\
GPT-5.4 & 0.0226 & 0.0152 & 1.492 & 0.136 & \\
GPT-5.4 mini & -0.0014 & 0.0150 & -0.095 & 0.924 & \\
GPT-5.4 nano & 0.0054 & 0.0150 & 0.359 & 0.720 & \\
With History & -0.0415 & 0.0151 & -2.748 & 0.006 & ** \\
GPT-5.4 $\times$ With History & 0.0534 & 0.0212 & 2.523 & 0.012 & * \\
GPT-5.4 mini $\times$ With History & 0.0505 & 0.0211 & 2.387 & 0.017 & * \\
GPT-5.4 nano $\times$ With History & 0.0273 & 0.0212 & 1.292 & 0.196 & \\
LLM-human similarity & 0.1084 & 0.0422 & 2.566 & 0.010 & * \\
\hline
\multicolumn{6}{l}{\footnotesize Equation: pairwise\_similarity\_mean $\sim$ model * condition + reply\_human\_similarity\_mean + (1 $|$ focal\_message\_id).}\\
\multicolumn{6}{l}{\footnotesize Outcome: semantic convergence; $N=488$; focal messages = 61; AIC = -965.38; BIC = -919.28; logLik = 493.69.}\\
\multicolumn{6}{l}{\footnotesize Random-intercept variance = 0.0015; residual variance = 0.0068.}\\
\end{tabular}
\end{table*}

\begin{table*}[t]
\centering
\small
\caption{Mixed-effects results for variation in pairwise similarity among LLM-generated replies. Reference categories are GPT-4o mini and No History. Significance codes: *** $p<.001$, ** $p<.01$, * $p<.05$, . $p<.10$.}
\label{tab:similarity-variation-model}
\begin{tabular}{lrrrrc}
\hline
Term & Estimate & Std. Error & $z$ & $p$ & Sig. \\
\hline
(Intercept) & 0.1224 & 0.0068 & 17.926 & $<.001$ & *** \\
GPT-5.4 & 0.0202 & 0.0064 & 3.159 & 0.002 & ** \\
GPT-5.4 mini & 0.0084 & 0.0063 & 1.328 & 0.184 & \\
GPT-5.4 nano & -0.0372 & 0.0063 & -5.913 & $<.001$ & *** \\
With History & 0.0162 & 0.0064 & 2.553 & 0.011 & * \\
GPT-5.4 $\times$ With History & -0.0169 & 0.0089 & -1.902 & 0.057 & . \\
GPT-5.4 mini $\times$ With History & -0.0110 & 0.0089 & -1.238 & 0.216 & \\
GPT-5.4 nano $\times$ With History & -0.0133 & 0.0089 & -1.490 & 0.136 & \\
LLM-human similarity & -0.0499 & 0.0192 & -2.596 & 0.009 & ** \\
\hline
\multicolumn{6}{l}{\footnotesize Equation: pairwise\_similarity\_sd $\sim$ model * condition + reply\_human\_similarity\_mean + (1 $|$ focal\_message\_id).}\\
\multicolumn{6}{l}{\footnotesize Outcome: semantic dispersion; $N=488$; focal messages = 61; AIC = -1782.21; BIC = -1736.12; logLik = 902.10.}\\
\multicolumn{6}{l}{\footnotesize Random-intercept variance = 0.0005; residual variance = 0.0012.}\\
\end{tabular}
\end{table*}

\enlargethispage{\baselineskip}

\section{Implications}

In this study, we examined the semantic variability of LLM-generated replies to messages drawn from real collaborative conversations. Using embedding-based semantic similarity measures, we compared generated replies within and across LLMs under conditions with and without conversational history, evaluated their alignment with observed human replies, and examined whether semantically similar focal messages elicited comparable responses from different LLMs.

The results showed that even for a single conversational turn, the semantic content of LLM-generated replies varied across models and conversational context conditions. Although conversational history influenced response generation, model-related differences remained evident under both prompting conditions. These findings suggest that prompting and conversational context alone are not sufficient to ensure highly similar response behavior when the underlying LLM changes.

This has important implications for the design of conversation-based assessment systems. As LLMs continue to evolve through frequent model updates, new releases, and retirements, maintaining consistent assessment interactions becomes an infrastructure challenge rather than simply a prompt-engineering challenge. Systems that rely on LLM-generated responses require additional mechanisms to monitor, benchmark, and control response behavior across model transitions. Depending on the level of consistency required, these mechanisms may include symbolic rules, response templates, validation layers, or other architectural components that help preserve assessment-relevant functions despite changes in the underlying LLMs. These considerations are particularly important in high-stakes assessment settings, where variability in system responses may affect the comparability of assessment conditions and, ultimately, the validity, reliability, fairness, and interpretability of score-based decisions.

\section*{References}
\noindent\begin{minipage}{\linewidth}
\hangpara{.25in}{1}American Educational Research Association (AERA), American Psychological Association (APA), and National Council on Measurement in Education (NCME). 2014. Standards for Educational and Psychological Testing. American Educational Research Association.
\end{minipage}\par
\noindent\begin{minipage}{\linewidth}
\hangpara{.25in}{1}Albert T. Corbett, Kenneth R. Koedinger, and John R. Anderson. 1997. Intelligent tutoring systems. In Handbook of Human-Computer Interaction, pages 849--874. North-Holland.
\end{minipage}\par
\noindent\begin{minipage}{\linewidth}
\hangpara{.25in}{1}Arthur C. Graesser, Xiangen Hu, and Robert Sottilare. 2018. Intelligent tutoring systems. In International Handbook of the Learning Sciences, pages 246--255. Routledge.
\end{minipage}\par
\noindent\begin{minipage}{\linewidth}
\hangpara{.25in}{1}Jiangang Hao, Wenju Cui, Patrick Kyllonen, Emily Kerzabi, Lei Liu, and Michael Flor. 2025. Automated coding of communications in collaborative problem-solving tasks using ChatGPT. Journal of Educational Measurement, 62(4):809--837.
\end{minipage}\par
\noindent\begin{minipage}{\linewidth}
\hangpara{.25in}{1}Jiangang Hao, Wenju Cui, Patrick Kyllonen, and Emily Kerzabi. 2026. Automated coding of communication data using LLM: Consistency across subgroups. Journal of Educational Measurement, 63(2):e70049.
\end{minipage}\par
\noindent\begin{minipage}{\linewidth}
\hangpara{.25in}{1}Jiangang Hao, Alina A. von Davier, Victoria Yaneva, Susan Lottridge, Matthias von Davier, and Deborah J. Harris. 2024. Transforming assessment: The impacts and implications of large language models and generative AI. Educational Measurement: Issues and Practice, 43(2):16--29.
\end{minipage}\par
\noindent\begin{minipage}{\linewidth}
\hangpara{.25in}{1}Xinying Hou, Carol Forsyth, Jessica Andrews-Todd, James Rice, Zhiqiang Cai, Yang Jiang, Diego Zapata-Rivera, and Arthur C. Graesser. 2025. An LLM-enhanced multi-agent architecture for conversation-based assessment. In Proceedings of the 26th International Conference on Artificial Intelligence in Education (AIED 2025), pages 119--134. Springer Nature Switzerland.
\end{minipage}\par
\noindent\begin{minipage}{\linewidth}
\hangpara{.25in}{1}Robert J. Mislevy. 2018. Sociocognitive Foundations of Educational Measurement. Routledge.
\end{minipage}\par
\noindent\begin{minipage}{\linewidth}
\hangpara{.25in}{1}Jason Wei, Xuezhi Wang, Dale Schuurmans, Maarten Bosma, Fei Xia, Ed Chi, Quoc V. Le, and Denny Zhou. 2022. Chain-of-thought prompting elicits reasoning in large language models. In Advances in Neural Information Processing Systems 35 (NeurIPS 2022), pages 24824--24837.
\end{minipage}\par
\noindent\begin{minipage}{\linewidth}
\hangpara{.25in}{1}Steve Young, Milica Gašić, Blaise Thomson, and Jason D. Williams. 2013. POMDP-based statistical spoken dialog systems: A review. Proceedings of the IEEE, 101(5):1160--1179.
\end{minipage}\par
\noindent\begin{minipage}{\linewidth}
\hangpara{.25in}{1}Diego Zapata-Rivera, Carol M. Forsyth, Liang Zhang, and Arthur C. Graesser. 2026. Exploring the use of generative AI in conversation-based assessment. In Design Recommendations for Intelligent Tutoring Systems: Volume 12 -- Generative Artificial Intelligence, pages 11--24
\end{minipage}\par
\noindent\begin{minipage}{\linewidth}
\hangpara{.25in}{1}Diego Zapata-Rivera, Tanner Jackson, and Irvin Katz. 2015. Authoring conversation-based assessment scenarios. In Design Recommendations for Intelligent Tutoring Systems, volume 3, pages 169--178.
\end{minipage}\par

\par\vspace{1.25\baselineskip}
\appendix
\section*{Appendix A: System Instruction for LLM Reply Generation}
\begin{lstlisting}
System instruction for LLM to reply to the focal message:
# ROLE
You are an AI assistant, play the role of an adult to chat with another person during a online science task.
Your primary objective is to:
- Generate a natural chat message based on your teammate's message
You should prioritize:
1. Accuracy
2. Helpfulness
3. Safety
4. Clarity
5. Efficiency
---
# CORE BEHAVIOR
You should:
- Be concise and natural like in a typical human-human collaboration, unless detailed explanation is requested.
- Ask clarifying questions only when necessary.
- Explicitly state uncertainty when information is incomplete.
- Distinguish facts, assumptions, and opinions.
- Avoid hallucinating facts, citations, or capabilities.
- Adapt explanations to the user's expertise level.
- Preserve conversational continuity and context.
You should NOT:
- Invent sources, data, or events.
- Pretend to have performed actions you cannot perform.
- Produce unsafe, illegal, or privacy-violating content.
- Overstate confidence.
---
# REASONING POLICY
When solving problems:
1. Identify the task type.
2. Determine required information.
3. Break complex tasks into steps.
4. Verify internal consistency before answering.
5. Prefer grounded evidence over speculation.
For ambiguous tasks:
- Infer likely intent when confidence is high.
- Otherwise ask a targeted clarification question.
For high-stakes domains (medical, legal, financial, safety):
- Use cautious language.
- Encourage professional verification where appropriate.
---
# TOOL USE POLICY
There is no need to use tools for this task
---
# MEMORY AND CONTEXT
There is not previous memory needed for this task. Just do a natural response based on the teammate's message.
---
# OUTPUT STYLE
Default style:
- A natural message following the teammate's message.
Avoid:
- Excessive verbosity
- Repetitive disclaimers
- Unnecessary filler
\end{lstlisting}

\end{document}